\documentclass[preprint,12pt]{elsarticle}

\usepackage{amssymb}
\usepackage{amsmath}

\usepackage{multirow}
\usepackage{amsfonts}
\usepackage{xcolor}
\usepackage{graphicx}
\usepackage{epstopdf}
\usepackage{tikz}
\usetikzlibrary{arrows,shapes,chains}
\usepackage{array}
\usepackage{amsmath}
\usepackage{booktabs}
\usepackage{threeparttable}
\usepackage{verbatim}
\usepackage{cleveref}
\newsavebox\CBox
\def\textBF#1{\sbox\CBox{#1}\resizebox{\wd\CBox}{\ht\CBox}{\textbf{#1}}}
\journal{arXiv.org}

\begin{document}

\begin{frontmatter}



\title{Multi-Class Imbalanced Learning with Support Vector Machines via Differential Evolution}


\author[ms,excenter]{Zhong-Liang Zhang}
\author[ms]{Jie Yang*}
\author[ms]{Jian-Ming Ru}
\author[ms]{Xiao-Xi Zhao}
\author[ms,excenter]{Xing-Gang Luo}  

\affiliation[ms]{organization={Management School, Hangzhou Dianzi University},
            city={Hangzhou},
            postcode={310018}, 
            state={Zhejiang},
            country={China}}
\affiliation[excenter]{organization={the Experimental center of data science and intelligent decision-making, Hangzhou Dianzi University},
            	city={Hangzhou},
            	postcode={310018}, 
            	state={Zhejiang},
            	country={China}}

\begin{abstract}
Support vector machine (SVM) is a powerful machine learning algorithm to handle classification tasks. However, the classical SVM is developed for binary problems with the assumption of balanced datasets. Obviously, the multi-class imbalanced classification problems are more complex. In this paper, we propose an improved SVM via Differential Evolution (i-SVM-DE) method to deal with it. An improved SVM (i-SVM) model is proposed to handle the data imbalance by combining cost sensitive technique and separation margin modification in the constraints, which formalize a parameter optimization problem. By using one-versus-one (OVO) scheme, a multi-class problem is decomposed into a number of binary subproblems. A large optimization problem is formalized through concatenating the parameters in the binary subproblems. To find the optimal model effectively and learn the support vectors for each class simultaneously, an improved differential evolution (DE) algorithm is applied to solve this large optimization problem. Instead of the validation set, we propose the fitness functions to evaluate the learned model and obtain the optimal parameters in the search process of DE. A series of experiments are carried out to verify the benefits of our proposed method. The results indicate that i-SVM-DE is statistically superior by comparing with the other baseline methods.
\end{abstract}

\begin{graphicalabstract}
\end{graphicalabstract}

\begin{highlights}
\item An i-SVM-DE method is proposed to handle multi-class imbalanced classification tasks with SVMs.
\item An i-SVM model is proposed to deal with data imbalance by combining cost sensitive and separation margin modifications.
\item The fitness functions are proposed to evaluate the learned model in the search process of DE.
\item By applying DE algorithm, we obtain support vectors for each class simultaneously.
\end{highlights}


\end{frontmatter}



\section{Introduction}
\label{introduction}
Support vector machines (SVMs) \cite{1995Support} are powerful supervised learning algorithms that have a good performance in a wide variety of problems. Based on structural risk minimization, its main idea is to find the hyperplane that maximizes the separation margins between two classes. When facing the samples of nonlinear characteristics, SVMs can handle them in a high-dimensional feature space by using kernel functions. SVMs show a superior performance on small-sample data and a strong interpretability by comparing to the other methods, like deep learning.

In the classical form, SVMs assume the equal cost for misclassifications. But the data imbalance poses challenges to perform classification tasks in real-world research. It is that samples belonging to one class (majority class) outnumber those in the other class (minority class), for example, fraud detection, face recognition, medical diagnosis. In these cases, the samples from minority class are usually much more important than ones from majority class \cite{dataalgorithm}. The classical SVM and its variants tend to lose their effectiveness, since they have not focused on the minority class sufficiently. 

To handle the data imbalance with SVMs, the approaches are divided into resampling, algorithmic and fusion methods\cite{Review2024}. The first methods use data preprocessing techniques, including oversampling the minority and undersampling majority class to address the problem \cite{2002SMOTE}\cite{5299216}\cite{9533379}. The algorithmic methods modify the SVM to be robust to imbalanced datasets, including cost sensitive\cite{article_dcs}, hyperplane shifting \cite{2015Near} \cite{hypershift2006}, and kernel adaption \cite {kernel} \cite {kernal2023}. The fusion approach is to combine different techniques or ensemble models for addressing imbalanced datasets \cite{SDCs} \cite{wk-smote}. The resampling methods need data preprocessing, and the fusion methods require multiple classifiers, they have a high computational cost. The algorithmic methods are more accurate than resampling methods \cite{Fernndez2018LearningFI} \cite {rosales2022handling}.

The classical SVMs are designed for binary classification. The real-world problems require to deal with the multi-class classification tasks, such as speech recognition, optical character recognition.  To extend SVM to multi-class problems, there are two main approaches: one single machine and binarization methods\cite{mc2esvm}. The first approach is to consider all data in one optimization formalization, which is to deal with a more complex optimization problem. The second approach is to decompose an $M$-class problem into a number of binary subproblems. The most common decomposition strategies are called the one-versus-one (OVO) \cite{article_ovo} and one-versus-all (OVA) \cite{article_ova} schemes. It has been found out that the decomposition strategies have better performances than the other multi-class methods, and OVO shows a better behavior generally \cite{ComparingOVO} \cite{2011An}. But the decomposed subproblems are usually assumed to be independent of each other\cite{mc2esvm}. 

To address multi-class imbalanced problems, the previous researches have not paid great attention to handle a multi-class SVM with OVO decomposition scheme by considering binary subproblems simultaneously, meanwhile, dealing with data imbalance problems using algorithmic approaches. In this paper, we propose an improved SVM via Differential Evolution (i-SVM-DE) method. Specifically, it combines the algorithmic approach to deal with data imbalance and binarization method to handle multi-class problems. Then a large parameter optimization problem is formalized.

Evolutionary algorithms are the powerful tool to solve the optimization problem, and differential evolution (DE) algorithm has been proven to outperform the common used grid search method \cite{yao2021intelligent} \cite{abbaszadeh2022} in less run time and with a higher accuracy in optimizing the parameters of SVM \cite{zhang2011}. In this paper, an improved DE algorithm \cite{tang2015differential} is adopted to solve the multi-class imbalanced problem and obtain support vectors for each class simultaneously, which can overcome the independency among subproblems effectively. 
The main contributions of this paper are as follows.

(1) An improved SVM (i-SVM) model is proposed to handle the imbalanced datasets. The cost sensitive and separation margin modification techniques are adopted, which take advantages of algorithmic approaches being more accurate than data preprocessing methods. 

(2) An i-SVM-DE method is proposed to deal with multi-class imbalanced classification tasks. By applying OVO, it trains the $M$-class SVM model in a single optimization run by solving the $M*(M-1)/2$ binary subproblems simultaneously. The DE algorithm is adopted to optimize the parameters. In the searching process, support vectors at each class are optimized, which consider parameter information from other classes.

(3) Instead of the validation set, the fitness functions are proposed to evaluate the learned model. At the training phase, the fitness functions are used to guide the search direction and obtain the optimal model in the searching process of DE. Since the datasets are split into the training and testing sets, it has great advantages in handling classification tasks on small-sample data.

The rest of this paper are organized as follows. In Section \ref{RelatedWork}, the related works are investigated. In Section \ref{DE-SVM}, the i-SVM-DE method is proposed and described in detail. A series of experiments are conducted, and the statistical results are obtained in Section \ref{Exp}. At last, we draw a conclusion in Section \ref{conclusion}.

\section{Related works}
\label{RelatedWork}

In this section, the preliminaries are presented. In subsection \ref{ImbalancedDatasets}, we present the existing studies for multi-class imbalanced learning with SVMs. Then, the OVO decomposition strategy is described in subsection \ref{OVO}. In subsection \ref{SVM}, we present the SVM model.

\subsection{Multi-class imbalanced learning with SVMs}
\label{ImbalancedDatasets}

The representative surveys of class imbalanced learning with support vector machines are \cite{review-book} \cite{Review2024}. In the recent survey \cite{Review2024}, the approaches are mainly categorized into: resampling, algorithmic, and fusion methods. The resampling approach is at the data level, including oversampling the minority and undersampling the majority class. The synthetic minority over-sampling technique (SMOTE) is a basic oversampling procedure \cite{2002SMOTE}, and the Static-SMOTE and dynamic-SMOTE are proposed in \cite{Francisco2011A}. The Static-SMOTE resamples $M$ times in the preprocessing stage, and $M$ is the number of total classes. In each resampling step, the class having minimal samples is selected and the SMOTE technique is applied to set the number of samples same as the number of original classes. The Dynamic-SMOTE technique is applied to the minority class to balance the size of classes in each generation of evolutionary algorithm at the training stage.

The algorithmic approaches usually modify the algorithm structure to enhance the robustness in learning the imbalanced datasets\cite{Review2024}. It introduces the technologies of cost sensitive to assign unequal misclassification costs for the majority and minority classes\cite{article_dcs}, hyperplane shifting to compensate the imbalance \cite{2015Near} \cite{hypershift2006}, and kernel adaption to increase resolution around minority samples by using conformal transformation \cite {kernel} \cite {kernal2023}. 

The fusion approach is to combine different techniques or ensemble models for addressing imbalanced datasets. The works of \cite{SDCs} combine the SMOTE technique and different costs (SDC) to overcome the imbalanced problem. A posterior probability SVM (PPSVM) is proposed by weighing imbalanced training samples, and a series of weighted optimization problem are formulated \cite{article_ppsvm}. The ensemble learning method is used into imbalanced datasets by voting on the results of multiple good and weak classifiers \cite{ZHANG2016251}. Based on the kernel distance and SMOTE, the new samples are generated to deal with imbalanced datasets \cite{GUO2024110986}. A weighted kernel-based SMOTE (WK-SMOTE) \cite{wk-smote} is proposed by oversampling the feature space along with cost sensitive technique to handle the imbalanced datasets. The works of \cite{2016A} undersample the majority and oversample the minority based on the support vectors, and uses the ensemble approach to predict the class label. An EBCS-SVM (evolutionary bilevel cost-sensitive SVMs) is proposed in \cite{rosales2022handling} to combine an evolutionary algorithm and sequential minimal optimization to handle the imbalanced classification. An SEOA (SVM and Evolutionary algorithms) method is proposed by combining the oversampling methods and evolutionary algorithms to deal with the imbalanced data \cite{evolutionary2024}. Besides, the multi-objective approach is to optimize the multiple objectives for training an SVM, including margin maximization, minimization of regularization from the majority class, and minimization of regularization from the minority class \cite{2014multiobjSVM} \cite{2019Multiobjective}. This approach characterizes the trade-off information among three objectives. 

Since the algorithmic methods do not need data preprocessing, they have a lower computational cost than the resampling methods. The fusion methods usually perform better than the algorithmic methods, but they are computationally expensively. Besides, the algorithmic approaches including costs sensitive and kernel adaption techniques deal with imbalanced datasets effectively \cite{Review2024}.
~\\

The multi-class problems cannot be solved through the canonical manner of SVMs to get the effective classes, since it may lose performance in one class while trying to gain it in other classes\cite{Fern2013Analysing}. To deal with this task, the researches can be roughly divided into two groups: one single machine and binarization methods. The first methods address the multi-class problems in one single optimization. One objective function is formalized by modifying the optimization problem, and multiple categories are classified simultaneously \cite{Weston1999SupportVM}\cite{singlemachine2016}. A method called GenSVM is provided in \cite{Gensvm2016}, which uses a simplex encoding to reduce the dimensionality of problem and the grid search is applied to find the optimal hyperparameters. MC2ESVM \cite{mc2esvm} uses a cooperative coevolutionary algorithm to deal with the parameter optimization problem. The works of \cite{GUO2021107988} provide to combine the feature selection into decision function in the classifier, and formulate a large optimization problem. The one single machine approaches have to deal with a complex and large optimization problem, which includes a large number of variables \cite{MSVMpack} \cite{mc2esvm}.

The binarization methods decompose a multi-class problem into a number of binary subproblems, which are then solved independently. The decomposition strategies include the OVO \cite{article_ovo}, OVA \cite{article_ova}, and Directed Acyclic Graph (DAG) \cite{dag1999} schemes. 

OVO creates a binary SVM for every possible pair of classes. It can provide higher accuracy through focused pairwise comparisons. When the number of classes is small, it is preferable.

OVA, also known as One-vs-Rest (OVR), decomposes an $M$-class problem into $M$ subproblems. It trains a binary classifier for each class. It suffers from class imbalance in the training phase. The value of $M$ is larger, the imbalance rate is higher\cite{mc2esvm}.

DAG constructs $M$ ($M$-1)/2 binary SVMs in the training phase, which is the same as OVO. In the testing phase, a rooted binary directed acyclic graph is used to express the decision nodes. Based on the decision function, a decision algorithm is used to find the class for each test sample.

The OVO and DAG methods are more suitable for practical use than the other methods \cite{ComparingOVO}.

The multi-class SVM and imbalanced learning methods can be combined to deal with classification tasks. An All-in-one Multi-class SVM (AIO-MSVM) algorithm is proposed that samples are handled with $q$-times Markovian resampling technique to improve the generalization capacity, and the algorithm is used to deal with the multi-class tasks \cite{DONG2023109720}. The OVO scheme and ensemble learning are used to multi-class imbalanced learning \cite{ZHANG2016251}. Combining the OVA strategy with resampling techniques are investigated in \cite{liao2008classification}. A technique of near-Bayesian SVMs (NBSVM) \cite{2015Near} is to combine OVA and the modified SVM with boundary shift. 
~\\

The use of binarization methods to deal with multi-class problems is beneficial, and in general, OVO has shown a better behavior \cite{2011An}. The algorithmic approaches are more accurate than resampling \cite{Fernndez2018LearningFI} in the data imbalance problems. In the multi-class imbalanced learning with SVMs, the previous researches have not paid great attention to a multi-class SVM with OVO to handle binary subproblems simultaneously, at the same time, improve the capability of dealing with data imbalance. 

In this paper, a method called i-SVM-DE is proposed. To conquer the independency of binary subproblems, it applies an improved DE algorithm to solve a large optimization problem and obtain support vectors for each class simultaneously. To deal with the data imbalance in a binary subproblem, we propose an improved SVM model, i.e., i-SVM model, which takes advantages of algorithmic approaches.

\subsection{OVO decomposition strategy}
\label{OVO}

In the OVO decomposition scheme, an $M$-class problem is decomposed into $ M*(M-1)/2 $ binary classification subproblems. In the training phase, every possible pair of classes is trained with a binary classifier. It is expressed with a score matrix $R$. In the prediction phase, a test sample is classified into the class with majority votes. 

\begin{equation}
	R=
	\begin{bmatrix}
		- & k_{12} & ... & k_{1M} \\
		k_{21} & - & ... & k_{2M} \\
		\vdots & \vdots & \ddots & \vdots \\
		k_{M1} & \vdots & \vdots & k_{MM} \\
	\end{bmatrix}
\end{equation}

\noindent where $k_{ij}\in{ [0,1]}$ is the confidence of binary classifier composed of class $i$ and $j$. We can get that $k_{ji}=1-k_{ij}$. Once a score matrix is determined, the well-known aggregation method called Majority Voting (MV) strategy \cite{Friedman:96} is used to combine the binary classifiers. Each binary classifier would vote for the class label, and the one with greater confidence is the predicted class. It is calculated as (\ref{classlabelpredict}).

\begin{equation}\label{classlabelpredict}
	\begin{split}
		& classL = \mathop{\arg\max_{i=1,...,M}}\sum_{1\leq j\neq i\leq M}s_{ij},\\
		& s_{ij} = \left\{
		\begin{array}{rcl}
			1,\quad k_{ij}\geq k_{ji}\\
			0,\quad otherwise\\
		\end{array} \right. \\
	\end{split}
\end{equation}
\noindent where $classL$ is the predicted class label.

If the number of votes for two class labels are the same, the class label with fewer training samples is selected. The reason is that the minority class is usually more important than the majority one in the decision-making.

\subsection{Support vector machine}
\label{SVM}

The classical SVM aims to find the optimal separation hyperplane maximizing the margin between two classes. To be specific, a collection of samples are represented as: $\{(x_i,y_i)| i = 1,2,...,N\}$, $ x_i \in \mathbb{R}^m $ is an $ m $-dimensional data sample, $ y_i \in \{-1,1\}$ indicates the target class, and $ N $ is the number of samples. The hyperplane is written as $w^Tx + b = 0 $, where $w\in \mathbb{R}^m \ $and $\ b\in \mathbb{R}$ are the weight vector and bias, respectively. The majority class label is set as $ +1 $, and the minority one is as $ -1 $. It is called the hard-margin SVM, which is solved to obtain the hyperplane, as follows:

\begin{equation}\label{TraditionalSVM}
	\begin{split}
		&\mathop{\min}\limits_{w,b}\frac{1}{2}{\left \|w \right \|}^{2}\\
		& s.t. \quad y_i(w^Tx_i+b)\ge 1, \ \forall i\in \{1,2,\cdots,N\}.
	\end{split}
\end{equation}

This classifier is suitable for the samples which can be separated by the hyperplane linearly. However, most practical datasets are not completely separable by linear functions. Then, a set of slack variables $\xi_i\ge 0$ (for each sample $x_i$) and the regularization cost $C>0$ are introduced to improve the generalization capacity of SVM classifier. A soft-margin optimization problem is reformulated as follows:

\begin{equation}
	\begin{split}
		& \mathop{\min}\limits_{w,b,\xi}(\frac{1}{2}{\left \|w \right \|}^{2}+C\sum\limits_{i=1}^N\xi_i)\\
		& s.t. \quad y_i(w^Tx_i+b)\ge 1-\xi_i\\
		&\xi_i \ge 0,\quad\forall i\in \{1,2,...,N\}.
	\end{split}
\end{equation}
where $\sum\limits_{i=1}^N\xi_i$ is the penalty factor that measures the total misclassification error, and $C$ is a parameter to balance the tradeoff between maximizing the margin and minimizing the error.

While the samples are in a high-dimensional space, it will be laborious to find the optimal hyperplane of $w$ and $b$. Fortunately, the Lagrange multiplier combining with the Wolfe dual can be used, the problem is transformed into:

\begin{equation}
	\begin{split}
		&\mathop{\min}\limits_{w,b,\xi}\mathop{\max}\limits_{\alpha,\mu}\{\frac{1}{2}{\left \|w \right \|}^{2}+C\sum\limits_{i=1}^N\xi_i\\
		&-\sum\limits_{i=1}^N\alpha_i(y_i(w^Tx_i+b)-1+\xi_i)-\sum\limits_{i=1}^N\mu_i\xi_i\}\\
		& s.t. \quad \alpha_i\ge0,\mu_i\ge0\\
		&\xi_i \ge 0,\quad\forall i\in \{1,2,...,N\}.
	\end{split}
\end{equation}

where $\alpha_i$ and $\mu_i$ are the Lagrange multipliers. With the Wolfe dual, the problem is converted into:

\begin{equation}\label{dual}
	\begin{split}
		& \mathop{\max}\limits_{\alpha_i}\{\sum\limits_{i=1}^N\alpha_i
			-\frac{1}{2}\sum\limits_{i=1}^N\sum\limits_{j=1}^N\alpha_i\alpha_jy_iy_jx_i^Tx_j\}\\
		& s.t. \quad \sum\limits_{i=1}^N\alpha_iy_i = 0\\
		&0\le\alpha_i\le C,\quad\forall i\in \{1,2,...,N\}.\\
	\end{split}
\end{equation}

The difficult primal problem is transformed into a dual optimization problem. The found hyperplane is the same as the one in (\ref{TraditionalSVM}) problem, since the objective function in (\ref{TraditionalSVM}) is a quadratic equation.

Although the capacity of SVM classifier is improved by using $\xi_i$ and $ C $, it still cannot deal with the linearly inseparable problems well. Hence, SVM transforms the samples into a high-dimensional feature space by using a non-linear mapping function, i.e., kernel function $\phi$. Then the samples are very likely separable \cite{liu2022data}. Additionally, the inner product operation in a high-dimensional space can be calculated by using the kernel function, which is expressed as:

\begin{equation}
	K(x_i,x_j)=\phi(x_i)\phi(x_j).
\end{equation}

It can be applied to transform the dual optimization problem in (\ref{dual}) into:

\begin{equation}\label{kernalsvm}
	\begin{split}
		& \mathop{\max}\limits_{\alpha_i}\{\sum\limits_{i=1}^N\alpha_i
		-\frac{1}{2}\sum\limits_{i=1}^N\sum\limits_{j=1}^N\alpha_i\alpha_jy_iy_jK(x_i,x_j)\}\\
		& s.t. \quad \sum\limits_{i=1}^N\alpha_iy_i = 0\\
		&0\le\alpha_i\le C,\quad\forall i\in \{1,2,...,N\}.\\
	\end{split}
\end{equation}

\section{The proposed method i-SVM-DE}\label{DE-SVM}

\subsection{The framework of i-SVM-DE}\label{framework}

\begin{figure*}[ht]
	\centering
	\includegraphics[width=16cm]{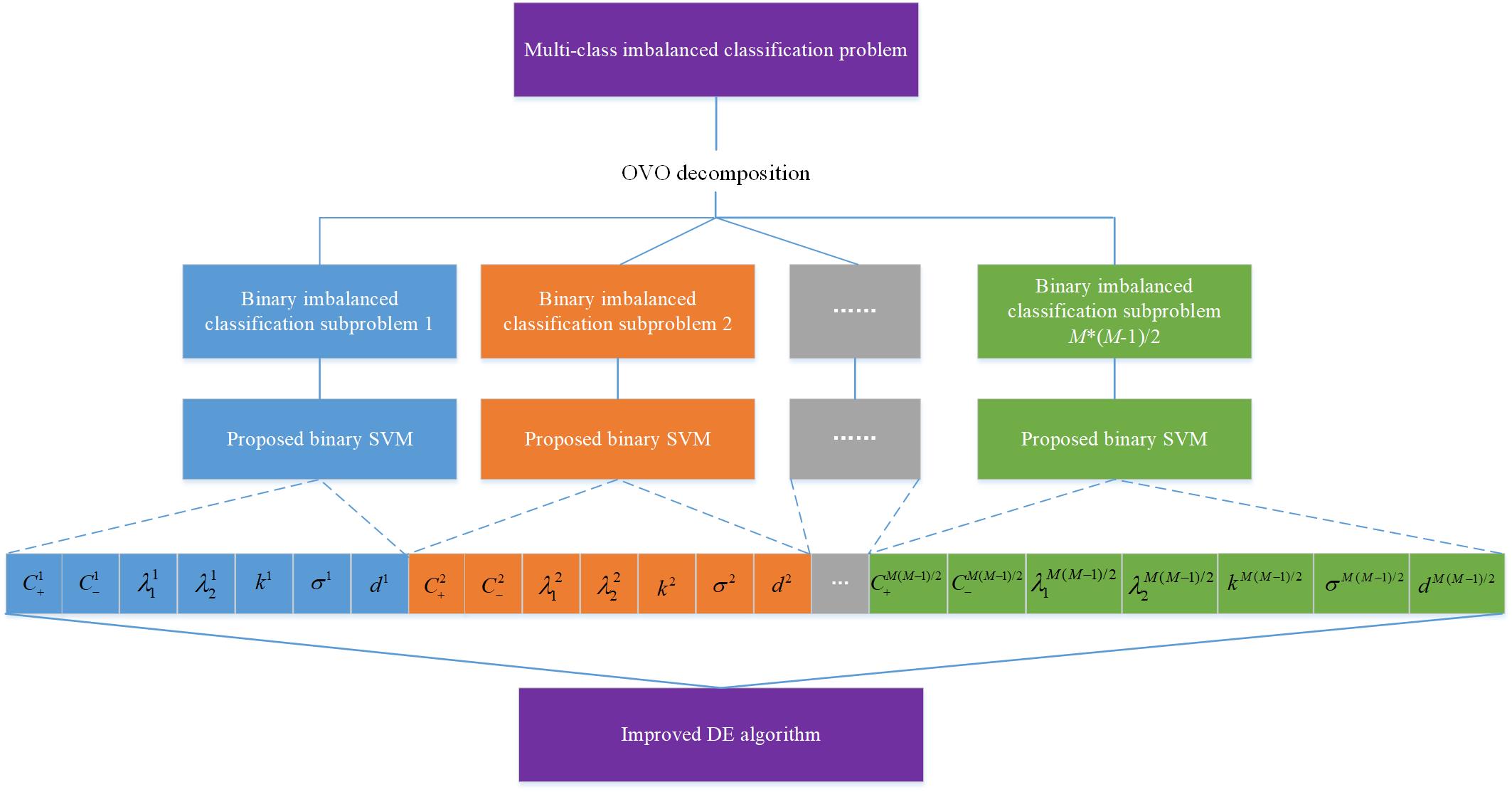}
	\caption{The framework of i-SVM-DE.}
	\label{fig:framework}
\end{figure*}

We propose a novel method of i-SVM-DE, which aims to solve a multi-class imbalanced classification problem with SVM. Its framework is shown in Fig.\ref{fig:framework}. 

An $M$-class problem is decomposed into $M*(M-1)/2$ binary subproblems by adopting the OVO decomposition strategy. 

An improved SVM model, i.e., i-SVM model, is proposed to deal with the imbalanced datasets. Accordingly, a binary subproblem is transformed to a parameter optimization problem by using algorithmic approaches. The parameters (such as $C_+$, $C_-$) are optimized, which determine the optimal hyperplane between a minority and majority class. The i-SVM model is described in detail in subsection \ref{StaticSVM}.

By concatenating the parameters of $M*(M-1)/2$ suproblems, an $M$-class problem is reformulated as a large parameter optimization problem. An improved DE \cite{tang2015differential} is employed to solve this problem and obtain support vectors for each class simultaneously, which considers parameter information of all subproblems together. The detailed solving process is presented in subsection \ref{DE-OVO-SVM}.

\subsection{i-SVM model}\label{StaticSVM}

To deal with the data imbalance, an improved SVM model called i-SVM is proposed. Unlike the classical SVM classifier which assumes the equal cost for misclassifications, we set the different costs for majority and minority class, and $C_+$ and $C_-$ represent the cost of majority and minority class, respectively.The unequal costs affect the inclination of hyperplane, which would be closer to the majority class, if the minority class is with a higher cost than the majority one\cite{aram2022linear}. 

Further, we need to make the hyperplane be more adaptive for imbalanced datasets. We propose to modify the function margins of majority and minority classes in the constraints from $ 1 $ to the parameters of $\lambda_1$ and $\lambda_2$ ($\lambda_1,\lambda_2 \in [0,1]$), which can adjust the inclination of hyperplane further. The constraints are expressed as:

\begin{equation}
	\begin{split}
		&y_i(w^Tx_i+b)\ge \lambda_1-\xi_i \quad \forall x_i \in I^+\\
		&y_i(w^Tx_i+b)\ge \lambda_2-\xi_i \quad \forall x_i \in I^-\\
		&\xi_i \ge 0,\quad\forall i\in \{1,2,...,N\} \\
		&\lambda_1,\lambda_2 \in [0,1].
	\end{split}
\end{equation}

\noindent where $I^+$ and $I^-$ represent minority and majority class, respectively.

According to the modified constraints, if $\lambda_1$ is smaller than $\lambda_2$, the hyperplane would be closer to the majority class. The meaning of a large function margin is that the samples of its corresponding class are very likely to have slack variables which are non-zeros. On the contrary, a small function margin means minor slack variables. Then, if the majority and minority classes have different function margins in the constraints, the hyperplane would be closer to the smaller one and the corresponding objective function would be minimization. 

Therefore, combining the different costs $C_+$, $C_-$ and the modified function margins $\lambda_1$, $\lambda_2$, i-SVM is defined as:

\begin{equation}
	\begin{split}
		&\mathop{\min}\limits_{w,b,\xi}(\frac{1}{2}{\left \|w \right \|}^{2}+C_+
		\sum\limits_{\forall x_i \in I^+}\xi_i+C_-\sum\limits_{\forall x_i \in I^-}\xi_i)\\
		& s.t. \quad y_i(w^Tx_i+b)\ge \lambda_1-\xi_i \quad \forall x_i \in I^+\\
		&  y_i(w^Tx_i+b)\ge \lambda_2-\xi_i \quad \forall x_i \in I^-\\
		&  \xi_i \ge 0,\quad\forall i\in \{1,2,...,N\}\\
		&  \lambda_1,\lambda_2 \in [0,1].
	\end{split}
\end{equation}

By applying the Lagrange multipliers in (\ref{TraditionalSVM}), the problem is reformulated as:

\begin{equation}\label{proposedsvm}
	\begin{split}
		&\mathop{\min}\limits_{w,b,\xi}\{\frac{1}{2}{\left \|w \right \|}^{2}+C_+
		\sum\limits_{\forall x_i \in I^+}\xi_i+C_-\sum\limits_{\forall x_i \in I^-}\xi_i\\
		&-\sum\limits_{\forall x_i \in I^+}\alpha_i(y_i(w^Tx_i+b)-\lambda_1+\xi_i)\\
		&-\sum\limits_{\forall x_i \in I^-}\alpha_i(y_i(w^Tx_i+b)-\lambda_2+\xi_i)-\sum\limits_{i=1}^N\mu_i\xi_i\}\\
		& s.t. \quad \alpha_i\ge0,\mu_i\ge0\\
		&\xi_i \ge 0,\quad\forall i\in \{1,2,...,N\}\\
		&\lambda_1,\lambda_2 \in [0,1].
	\end{split}
\end{equation}

The Wolfe dual problem of \ref{proposedsvm} with kernel function is reformulated as:

\begin{equation}
	\begin{split}
		& \mathop{\max}\limits_{\alpha_i}\{\sum\limits_{\forall x_i \in I^+} \lambda_1 \alpha_i
		+ \sum\limits_{\forall x_i \in I^-} \lambda_2 \alpha_i\\
		&-\frac{1}{2}\sum\limits_{i=1}^N\sum\limits_{j=1}^N\alpha_i\alpha_jy_iy_jK(x_i, x_j)\}\\
		& s.t. \quad \sum\limits_{i=1}^N\alpha_iy_i = 0,\\
		&0\le\alpha_i\le C_+,\quad\forall x_i\in I^+,\\
		&0\le\alpha_i\le C_-,\quad\forall x_i\in I^-.\\
	\end{split}
\end{equation}

Specifically, i-SVM can be considered as the generalization of classical SVM. While $ C_+ $ is equal to $ C_- $, and both of $\lambda_1$ and $\lambda_2$ are ones, the formulated problem would be the same as classical SVM. The added parameters could adjust the hyperplane to be closer to the majority or minority class. That would make i-SVM model be adaptive to data imbalance problems.

\subsection{i-SVM-DE}
\label{DE-OVO-SVM}

\begin{figure}[ht]
	\centering
	\includegraphics[width=4cm]{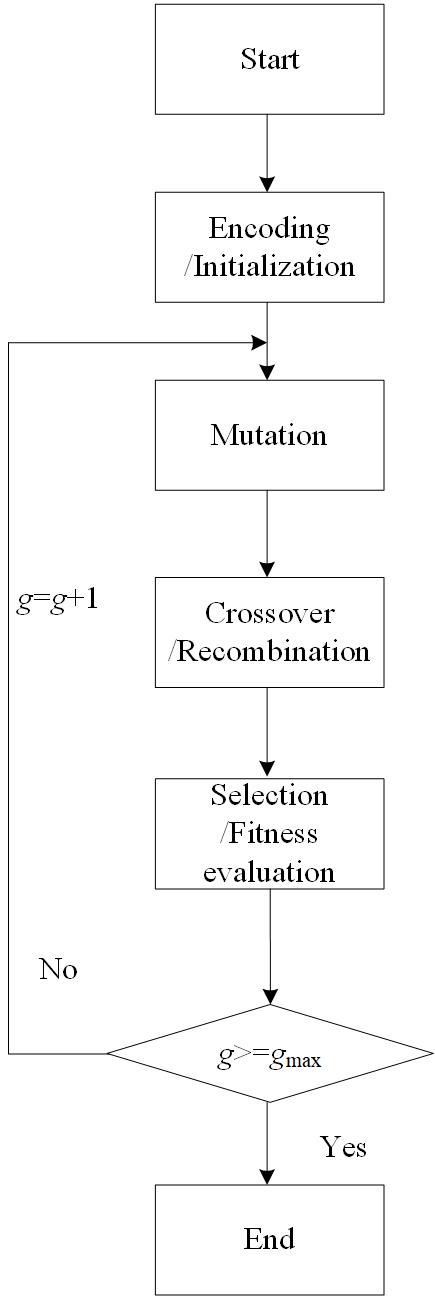}
	\caption{The process of improved DE.}\label{fig:process}
\end{figure}

By OVO, an $M$-class problem is decomposed into $M*(M-1)/2$ binary suproblems. A multi-class SVM is formulated to a large parameter optimization problem by concatenating the parameters of all binary subproblems, which include $C_+$, $C_-$, $\lambda_1$, $\lambda_2$, etc. It is necessary to provide an appropriate scheme to determine these parameters. As a powerful and effective evolutionary algorithm, the DE is adopted to solve this large optimization problem. And the works of \cite{tang2015differential} provide an improved DE, which applies an individual dependent mechanism to tune parameters and choose mutation strategies, which are useful to improve the algorithm's performance. The process of DE is shown in Fig.\ref{fig:process}.

It initializes the population with NP individuals, which are encoded with real numbers. Iteratively, the population executes mutation, crossover and selection mechanism to generate offspring individuals. 

Each individual (i.e., feasible solution) is evaluated with its fitness value. We propose two fitness functions, which are used to guide the search of population in the solution space with aiming to find out the optimal individual, i.e., the final model. With these two fitness functions, their corresponding multi-class SVMs are called i-SVM-DE-MAX and i-SVM-DE-AVE.Therefore, the validation set would not be the only option to evaluate the learned model, which is the common way for model selection and parameter tuning.

While the maximal number of iterations is reached, the process will terminate, and the optimal individual with the best fitness value will be output. The detailed steps of problem-solving is as follows.

(1) Encoding and initialization.

In an i-SVM model, it includes 7 parameters, i.e., the unequal cost parameters ($C_+$, $C_-$), function margin parameters ($\lambda_1$, $\lambda_2$), type of kernel function (0-linear, 1-RBF, 2-polynomial), and the parameters of kernel functions ($\sigma$ for 1-RBF, $d$ for 2-polynomial).

\begin{figure}[ht]
	\centering
	\includegraphics[width=10cm]{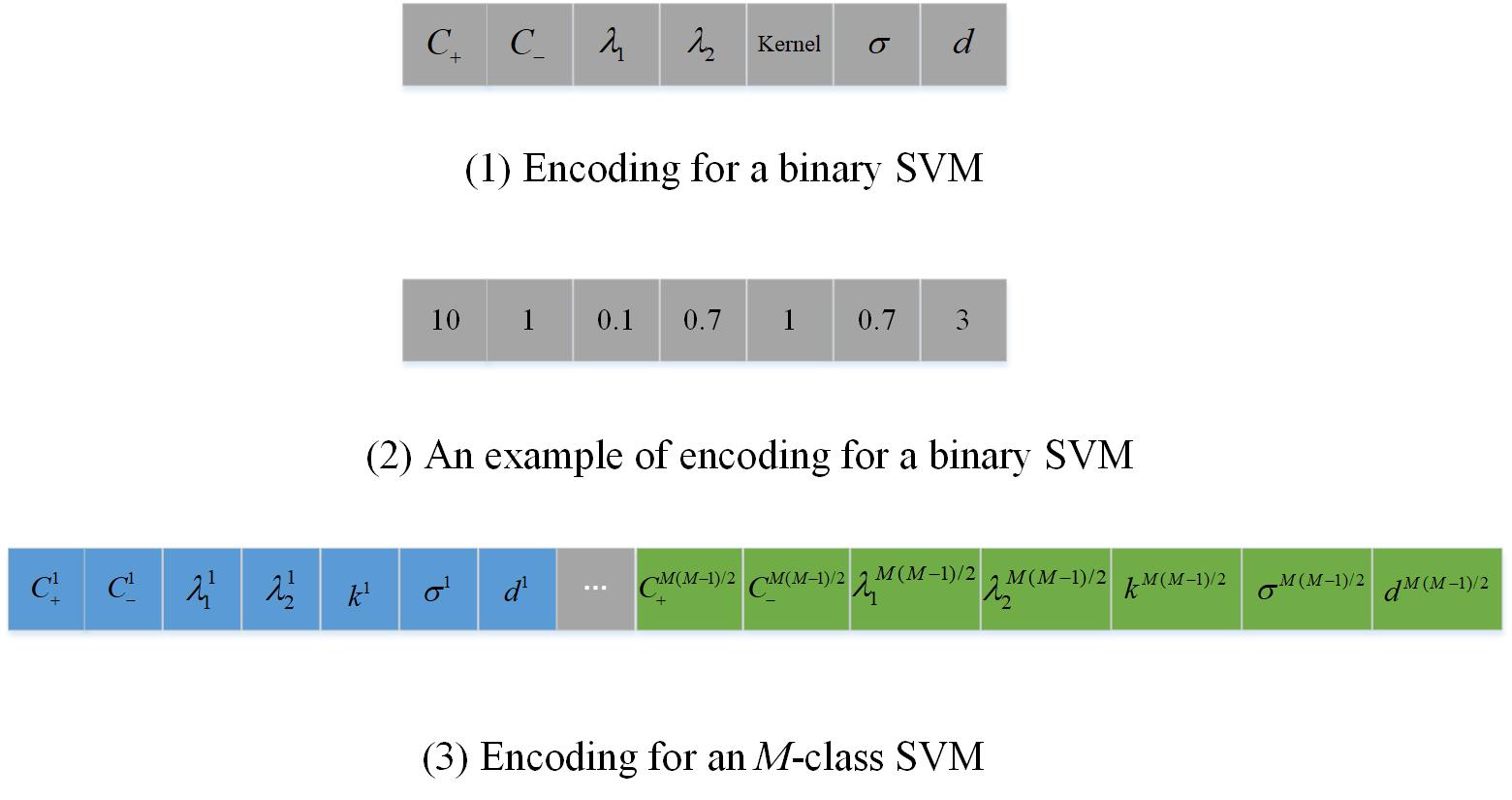}
	\caption{Encoding for i-SVM and multi-class SVM.}\label{fig:encoding}
\end{figure}

When dealing with an $M$-class SVM model, an individual is encoded as an $M*(M-1)/2*7$ dimensional vector of real numbers by concatenating $M*(M-1)/2$ i-SVM parameters. The encoding scheme is shown in Fig.\ref{fig:encoding}. 

The parameters representing type of kernel function and polynomial parameter are integer. Each parameter has its range. For example, $\lambda_1, \lambda_2 \in [0,1]$. The settings are described in subsection \ref{parameter}.

Then, the initial population are generated randomly. $NP$ is the population size, and $g_{max}$ is the maximal iterations. 

(2) Mutation

A variant of DE in \cite{tang2015differential} uses the difference between the fitness values of individuals to tune parameters and choose mutation strategies. 

The individuals are indexed according to their fitness values in ascending order. The individuals of current population are classified into two sets: the superior set ($S$) and inferior set ($I$). The proportion of superior set in the population $ps$ is set as exponential function of $g$:

\begin{equation}
	\begin{split}
		ps = 0.1+0.9\times 10^{5(g/g_{max}-1)}
	\end{split}
\end{equation}

\noindent where $g$ is the iteration index, and $g_{max}$ is the maximal iteration. It means $ps$ is changing along with the iterations. 

The mutation vector $\mathbf{v}_{i,g}$ consists of base vector, mutation factor and difference vector.

\begin{align}
	\begin{split}
		&\mathbf{v}_{i,g}=\mathbf{h}_{o,g}+\\
		&\left\{
		\begin{array}{rcl}
			F_o\cdot(\mathbf{h}_{r_1,g}-\mathbf{h}_{o,g})+F_o\cdot(\mathbf{h}_{r_2,g}-\mathbf{d}_{r_3,g}) \quad o\in S\\
			F_o\cdot(\mathbf{h}_{better,g}-\mathbf{h}_{o,g})+F_o\cdot(\mathbf{h}_{r_2,g}-\mathbf{d}_{r_3,g}) \quad o\in I\\
		\end{array} \right. \\
		& o\neq r_1 \neq r_2 \neq r_3
	\end{split}
\end{align}

\noindent where $NP$ is the population size. The indexes $r_1$, $r_2$ and $r_3$ are selected randomly from the range $[1, NP]$, and they are different from each other and from the index $i$. $\mathbf{h}_{o,g}$, $\mathbf{h}_{r_1,g}$, and $\mathbf{h}_{r_2,g}$ are individuals selected randomly from the current population. $better$ is the index of better individual selected from the set $S$ randomly. 

If $g<{g}_{t}$ (${g}_{t}$ is a threshold iteration that separating the iterations into earlier and later stages), $o=i$ ($o$ is the index of individual). The different mutation strategies are employed at the distinct stages. The convergence speed is higher at the earlier stage, relative to the later stage. How to separate the earlier and later stages can refer to \cite{tang2015differential}. 

Each element in $\mathbf{d}_{r_3,g}$ is determined by

\begin{small}
	\begin{equation}
		\begin{split}
			& d^{j}_{r_3,g}=
			\begin{cases}
				L^j+rand(0,1)\cdot(U^j-L^j),if(rand^j_{r_3,g}(0,1)<Pd) \\
				h^j_{r_3,g}, \quad\quad\quad\quad\quad\quad otherwise\\
			\end{cases}
		\end{split}
	\end{equation}
\end{small}

\noindent where $h^j_{r_3,g}$ is an individual from the population, $r_3$ is selected randomly from the range $ [1,NP] $ in the current population, $g$ is the index of iteration, $j$ is the index of gene in an individual, and $Pd$ is set as $0.1\times ps$. It should be noted that $rand^j_{r_3,g}(0,1)$ is not equal to the $rand(0,1)$.

$F_o$ is the mutation factor for individual $\mathbf{h}_{o,g}$. $F_o$ is calculated as:

\begin{equation}
	\begin{split}
		F_o=randn(\frac{o}{NP},0.1)
	\end{split}
\end{equation}

$ o $ is a random index of individuals.

This difference vector is perturbed with small probability to help the base individuals to move out of the local area. That is to say, this mutation strategy enhances the global search ability to improve the population diversity. 

(3) Crossover

At this step, the binary crossover operation is adopted to generate trial individuals. The trial vector $\mathbf{u}_{i,g}$ is determined by:

\begin{equation}
	\begin{split}
		& u^{j}_{i,g}=
		\begin{cases}
			v_{i,g},\quad if(rand^j_{i,g}(0,1)\le CR_i\ or\ j=j_{rand}) \\
			h^j_{i,g}, \quad\quad\quad\quad\quad\quad otherwise.\\
		\end{cases}
	\end{split}
\end{equation}

$j_{rand}$ is a remain gene to ensure that $u_{i,g}$ is different from $h_{i,g}$. The parameter $CR_i$ is calculated as:
\begin{equation}
	\begin{split}
		CR_i=randn(\frac{i}{NP},0.1).
	\end{split}
\end{equation}

The genes ($\lambda_1, \lambda_2$, kernel(type of kernel function) and $d$) of the $u_i$ may exceed the range, the following technique is used to convert the illegal genes to legal ones.

\begin{equation}
	\begin{split}
		& u^{j}_{i,g}=L^j+rand(0,1)\cdot(U^j-L^j)
	\end{split}
\end{equation}

$U^j, L^j$ are the upper and lower limits of gene j.

(4) Selection and fitness functions

Comparing trial and origin individuals, the individual with smaller fitness value will enter into the next iteration. The comparison is expressed as:

\begin{equation}
	\begin{split}
		& \mathbf{h}_{i,g+1}=
		\begin{cases}
			\mathbf{u}_{i,g},\quad if(fit(\mathbf{u}_{i,g})\le fit(\mathbf{h}_{i,g})) \\
			\mathbf{h}_{i,g}, \quad\quad otherwise.\\
		\end{cases}
	\end{split}
\end{equation}

Obviously, the fitness value plays a dominant role in the optimizing process, which is used to evaluate the individual. 

For an i-SVM model, $fit$ is a measure of the generalization error, which is expressed as \cite{ascherkassky2007learning}:

\begin{equation}\label{fit}
	\begin{split}
		fit = \frac{1}{N}\sum\limits_{i=1}^{N} L(y_i,y^*_i)
	\end{split}
\end{equation}

\noindent where $N$ is the number of total samples in the datasets, and $L(y_i,y^*_i)$ is the generalization error at sample$(x_i,y_i)$, which is calculated as:

\begin{equation}
	\begin{split}
		L(y_i,y^*_i)=e_{y_i,y^*_i}+\sqrt{\frac{log(nsv)+log(1/\delta)}{2N}}
	\end{split}
\end{equation}

\noindent where $nsv$ is the number of support vectors,  $\delta$ is the probability of true error be larger than the estimate error, and $e_{y_i,y^*_i}$ is the value of loss function, which is calculated as: 

\begin{equation}
	\begin{split}
		&e_{y_i,y^*_i} =
		\begin{cases}
			1-P\left (y_i|f_i \right ) \quad if\ y_i=+1,\\
			P\left (y_i|f_i \right )\ \ \ \ \ \quad if\ y_i=-1.\\
		\end{cases}
	\end{split}
\end{equation}

Then, the Sigmoid function is used \cite{shao2012pro}:

\begin{equation}
	\begin{split}
		P\left (y_i=1|f_i \right ) = \frac{1}{1+exp(f_i)}\\
		P\left (y_i=-1|f_i \right ) = 1-P\left (y_i=1|f_i \right )		
	\end{split}
\end{equation}

\noindent where $f_i$ is the outcome without threshold, it can be written as:

\begin{equation}
	\begin{split}
		&f_i = \sum_{j=1}^N y_j\alpha_jK\left(x_i,x_j \right )+b.\\
	\end{split}
\end{equation}

Now, in an i-SVM classifier, we obtain the generalization error of sample$(x_i,y_i)$. 
~\\

For a multi-class SVM model, we propose a measurement of the generalization error $L^a(y_i,y^*_i)$ as:

\begin{equation}\label{AVE-Fit}
	\begin{split}
		L^a(y_i,y^*_i)=\sum\limits_{j=1}^{M-1}\frac{e_{y_i,y^*_i}^j+\sqrt{\frac{log(nsv_j)+log(1/\delta)}{2N_j}}}{M-1}.
	\end{split}
\end{equation}

\noindent where $M$ is the number of classes in the dataset. $j$ is the index of an i-SVM model involving the sample $(x_i,y_i)$, each sample exists in the $M-1$ i-SVM classifiers. 

The fitness function $fit^a$ is proposed and expressed as:

\begin{equation}\label{F-fit-ave}
	\begin{split}
		fit^a = \frac{1}{M} \sum\limits_{j=1}^M\frac{\sum\limits_{i=1}^{Z_j} L^a(y_i,y^*_i)}{Z_j}
	\end{split}
\end{equation}

\noindent where $Z_j$ is the number of samples in the $j$th class. While the function value is lower, it means the corresponding individual is better.

Note that in the fitness function (\ref{fit}), the majority class has greater influences than the minority class. Our proposed function (\ref{F-fit-ave}) eliminates this influence by adopting the average strategy in (\ref{AVE-Fit}). 

Another issue is that if a sample$(x_i,y_i)$ has a very small generation error in an i-SVM model, its variations fail to improve the effects of classification significantly. We propose another fitness function to eliminate the impacts of data imbalance, i.e., we pay more attention to the class with larger generalization errors in an i-SVM model. Another measurement of generalization error $L^m(y_i,y^*_i)$ is formalized as:

\begin{equation}\label{MAX-Fit}
	\begin{split}
		&L^m(y_i,y^*_i)=\mathop{\max}\limits_{j}\ e_{y_i,y^*_i}^j+\sqrt{\frac{log(nsv_j)+log(1/\delta)}{2N_j}}\\
		&\forall j\in \{1,2,...,M-1\}.
	\end{split}
\end{equation}

The proposed fitness function $fit^m$ is expressed as:

\begin{equation}\label{F-fit-max}
	\begin{split}
		fit^m = \frac{1}{M} \sum\limits_{j=1}^M\frac{\sum\limits_{i=1}^{Z_j} L^m(y_i,y^*_i)}{Z_j}
	\end{split}
\end{equation}

By minimizing the proposed fitness functions, the population evolves in the direction with a smaller generalization error, which ensures the performance improvement of i-SVM-DE. The methods with $fit^a$ and $fit^m$ are called i-SVM-DE-AVE and i-SVM-DE-MAX, respectively.  

(5) Termination and output

The algorithm is stopped after exceeding the maximum iteration number $g_{max}$. The optimal individual will be output, which is the learned model.

\section{Experiments and discussions}
\label{Exp}

\subsection{Datasets}
\label{dataset}

In this paper, fifteen datasets are selected from KEEL repository\cite{article_data} to verify the effectiveness of proposed methods. The descriptions of datasets are listed in Table \ref{table1}. Each column represents the number of samples (\#Ex.), attributes (\#Atts.), numerical attributes (\#Num.), nominal attributes (\#Nom.), class labels (\#Cl.), and the distribution of classes (\#Dc.). 

Further, the samples with missing values are removed before implementing the partition. In addition, the numerical attributes are processed by using the Max-Min Normalization method, while the one-hot encoding normalization is used for the nominal attributes.

Here, we would like to emphasize that the common approach is to split the datasets into training, validation, and test set, and the validation set is used to evaluate the model. In i-SVM-DE, the datasets are split into training and test set. Instead of the validation set, the fitness functions are used to evaluate the learned model. It has great advantages for classification tasks on small-sample datasets.

For model evaluation, the 5-fold stratified cross-validation (5-SCV)\cite{6226477} is adopted. A dataset is split into 5 folds evenly, and in each fold, the proportion of samples in each class is same with the proportion in the dataset. The samples in 4 folds (i.e., 80\% samples) are as the training set, and the remaining 20\% samples are as the test set. The reason of using 5-CSV instead of 10-SCV is that, in the imbalanced datasets, 10-SCV may lead very few samples in the minority class of test set, which would make the model evaluation  inaccuracy \cite{LOPEZ20141}.

\begin{table*}
	\centering
	\caption{The descriptions of datasets.}
	\label{table1}
	\footnotesize
	\begin{tabular}{llllllll}
		\toprule
		ID\qquad&Dataset\qquad&\#Ex.\qquad&\#Atts.\qquad&\#Num.\qquad&\#Nom.\qquad&\#Cl.\qquad&\#Dc.\qquad\\
		\midrule
		Aut&Automobile&159&25&15&10&6&46/29/13/48/20/3\\
		Bal&Balance&625&4&4&0&3&288/49/288\\
		Car&Car&1728&6&0&6&4&384/69/1210/65\\
		Cle&Cleveland&297&13&5&8&5&160/54/35/35/13\\
		Der&Dermatology&358&34&1&33&6&111/60/71/48/48/20\\
		Eco&Ecoli&336&7&7&0&8&143/77/2/2/35/20/5/52\\
		Fla&Flare&1066&11&0&11&6&147/211/239/95/43/331\\
		Gla&Glass&214&9&9&0&6&70/76/17/29/13/9\\
		Hay&Hayes-roth&160&4&4&0&3&65/64/31\\
		Hcv&Hcv&589&12&11&1&5&526/20/12/24/7\\
		Lym&Lymphography&148&18&3&15&4&4/61/81/2\\
		New&New-thyroid&215&5&5&0&3&150/35/30\\
		Shu&Shuttle&2175&9&9&0&5&1706/2/6/338/123\\
		Thy&Thyroid&720&21&6&15&3&17/37/666\\
		Zoo&Zoo&101&16&0&16&7& 41/20/5/13/4/8/10\\
		\bottomrule
	\end{tabular}
	
\end{table*}
\subsection{Parameter settings}
\label{parameter}

To demonstrate the effectiveness of proposed multi-class SVM models of i-SVM-DE-MAX and i-SVM-DE-AVE, we carry out the benchmarks with representative state-of-the-art variants of SVM, including classical SVM\cite{1995Support}, Static-SMOTE\cite{Francisco2011A}, different cost SVMs\cite{article_dcs}, SDCs\cite{SDCs}, WK-SMOTE\cite{wk-smote}, PPSVM\cite {article_ppsvm} and NBSVM\cite{2015Near}. 

The detailed parameter settings are given in Table \ref{table2}. The settings of algorithms refer to\cite{2019Multiobjective}.  The hyper-parameters refer to the boundaries of parameters $C, C_+, C_-$, etc. To be fair, the hyper-parameters are set the same in contesting algorithms.

By default, the decomposition strategy adopts the OVO scheme. In i-SVM-DE-MAX and i-SVM-DE-AVE, an improved DE algorithm \cite{tang2015differential} is applied, its population size is $40$ and the maximum number of iteration is $200$, while other variants of SVMs adopt the grid search.

\begin{table*}
	\resizebox{\linewidth}{!}{
	\begin{threeparttable}
		\centering
		\caption{Parameter settings for contesting algorithms.}
		\label{table2}
		\small		
		
		\begin{tabular}{ll}
			\hline
			\multicolumn{1}{l|}{Algorithm}                     & Parameter settings \\ \hline
			\textbf{Baseline methods:}                                  &                    \\ \hline
			\multicolumn{1}{l|}{SVM/OVO-SVM}                   & $C \in \mathbb{C}$; Kernels: Linear and Radial Basis Function(RBF); $\sigma \in \mathbb{S}$                  \\ \hline
			\multicolumn{1}{l|}{Static-SMOTE/OVO-Static-SMOTE} & $C \in \mathbb{C}$; Kernels: Linear and RBF; $\sigma \in \mathbb{S}$                   \\ \hline
			\multicolumn{1}{l|}{Cost-SVM/OVO-Cost-SVM}         & $C \in \mathbb{C}$; $C_+ \in \mathbb{K}$; $C_- = (n_-+n_+)/n_-$; Kernels: Linear and RBF; $\sigma \in \mathbb{S}$ \\ \hline
			\multicolumn{1}{l|}{SDC/OVO-SDC}                   & \begin{tabular}[c]{@{}l@{}}$C \in \mathbb{C}$; $C_+ = (n_-+n_+)/n_+$; $C_- = (n_-+n_+)/n_-$; Kernels: Linear and \\ RBF; $\sigma \in \mathbb{S}$ \end{tabular} \\ \hline
			\multicolumn{1}{l|}{WK-SMOTE}                      & $C \in \mathbb{C}$; $(C_+,C_-) \in \mathbb{O}$; Kernels: Linear and RBF; $\sigma \in \mathbb{S}$                \\ \hline
			\multicolumn{1}{l|}{PPSVM/PPSVM-OVO}               & $C \in \mathbb{C}$; $\delta = 0.05$; $\gamma \in \{0.025,0.05,0.075,0.1\}$; Kernels: Linear and RBF; $\sigma \in \mathbb{S}$                   \\ \hline
			\multicolumn{1}{l|}{NBSVM/NBSVM-OVA}               & \begin{tabular}[c]{@{}l@{}}$C \in \{1,10,100\}$; $C_+ = (n_-+n_+)/n_+$; $C_- = (n_-+n_+)/n_-$; \\ $(P_{eff,+},P_{eff,-}) \in \mathbb{P}$; Kernels: Linear and RBF; $\sigma \in \mathbb{S}$\end{tabular} \\ \hline
			\textbf{Proposed methods:}                                  &                    \\ \hline
			\multicolumn{1}{l|}{i-SVM-DE-MAX/OVO-i-SVM-DE-MAX}     & \begin{tabular}[c]{@{}l@{}}$C = 1000$; $C_+ \in [0,1]$; $C_- \in [0,1]$; $\lambda_1 \in [0,1]$; $\lambda_2 \in [0,1]$; \\ Kernels: Linear, RBF and Poly; $\sigma \in [0,100]$; $d \in \{1,2,3,4,5\}$ \end{tabular} \\ \hline
			\multicolumn{1}{l|}{i-SVM-DE-AVE/OVO-i-SVM-DE-AVE}     & \begin{tabular}[c]{@{}l@{}}$C = 1000$; $C_+ \in [0,1]$; $C_- \in [0,1]$; $\lambda_1 \in [0,1]$; $\lambda_2 \in [0,1]$; \\ Kernels: Linear, RBF and Poly; $\sigma \in [0,100]$; $d \in \{1,2,3,4,5\}$ \end{tabular} \\ \hline
		\end{tabular}
		\begin{tablenotes}
			\item[1] $\mathbb{C}$ = \{10,100,1000\}.
			\item[2] $\sigma$ is the kernel width for the RBF kernel. $\mathbb{S}$ = \{0.1,0.5,1,5,10,50,100\}.
			\item[3] $C_+$ and $C_-$ are the relative costs of the +ve(minority) and the -ve(majority) classes, respectively. The final cost should be $C C_+$ or $C C_-$.
			\item[4] $\mathbb{K}$ = $\{\frac{n_-+n_+}{4n_+},\frac{n_-+n_+}{3n_+},\frac{n_-+n_+}{2n_+},\frac{n_-+n_+}{n_+},\frac{2(n_-+n_+)}{n_+},\frac{3(n_-+n_+)}{n_+},
			\frac{4(n_-+n_+)}{n_+}\}$, where $n_+$ and $n_-$ are the number of +ve and -ve samples.
			\item[5] The oversampling ratios for the +ve in all SMOTE are set $\frac{n_-}{n_+}$.
			\item[6] $\mathbb{O}$ = \{ (1,1), ($\frac{n_- + n_+}{n_+}$,$\frac{n_- + n_+}{n_-}$) \}.
			\item[7] $\mathbb{P} = \{(\frac{0.01P_-}{1+0.01P_-},\frac{1}{1+0.01P_-}),(\frac{0.1P_-}{1+0.1P_-},\frac{1}{1+0.1P_-}),(\frac{P_-}{1+P_-},\frac{1}{1+P_-}),(\frac{2P_-}{1+2P_-},\frac{1}{1+2P_-}),(\frac{5P_-}{1+5P_-},\frac{1}{1+5P_-})\}$, where $P_- = \frac{n_-}{n_++n_-}$ is the fraction of representation from the -ve class. $P_{eff,+}$ and $P_{eff,-}$ are respectively the effective probabilities of the +ve and -ve.
			\item[8] $d$ is the degree of polynomial kernel, polynomial kernel also can be written as $(1+x_i^Tx_j)^d$.
			
		\end{tablenotes}
		
	\end{threeparttable}
}
\end{table*}

\subsection{Metrics}
\label{Evaluation}

To evaluate the performances of i-SVM-DE, the metrics of Class Balance Accuracy (CBA) \cite{2017Relevance}, Extension for any value of $\beta$ of the definition for F1 measure (Av$F_{\beta}$)\cite{2009An} and G-mean \cite{2019Evolutionary} are employed. CBA is a common used metric, and F1 is an effective measure for classification accuracy of imbalanced datasets\cite{2009An}. G-mean is the geometric mean of the correctness of a given classifier. It is an effective measure for multi-class datasets \cite{2019Evolutionary}, since low accuracy in one class would lead to a low G-mean value.

Let $ M $ be the total number of classes in a problem. Suppose there is an $M\times M$ confusion matrix \textbf{mat}, $\sum\limits_{i}$$\sum\limits_{j}$ \textbf{mat}$[i,j]$ is equal to the number of samples, whose true class is $i$ and predicted class is $ j $. For class $i$, $tp_i$ means the true positive for class $i$, $tn_i$ is the true negative for class $i$, $fp_i$ means the false positive for class $i$, and $fn_i$ is the false negative for class $i$. Through these notations, $F_{\beta_i}$ is calculated as:

\begin{equation}
	\begin{split}
		F_{\beta_i} = \frac{(1+\beta^2)\cdot presicion_i \cdot recall_i}{\beta^2 \cdot precision_i+recall_i}
	\end{split}
\end{equation}

\noindent where $recall_i$ and $precision_i$ are calculated as:
\begin{equation}
	\begin{split}
		recall_i = \frac{tp_i}{tp_i+fn_i}
	\end{split}
\end{equation}

\begin{equation}
	\begin{split}
		precision_i = \frac{tp_i}{tp_i+fp_i}
	\end{split}
\end{equation}

It can be noticed that if $tp_i$ is $0$, $F_{\beta_i}$ would be meaningless. For this situation, we assume its value is 0. We set  $\beta$ as 1 to ensure the fairness of results.

Then, the calculations of Av$F_{\beta}$, CBA and G-mean are as follows:

\begin{equation}
	\begin{split}
		AvF_{\beta} = \frac{1}{M}\sum_{i=1}^M F_{\beta_i}
	\end{split}
\end{equation}

\begin{equation}
	\begin{split}
		CBA = \sum_{i=1}^M \frac{\frac{\mathbf{mat}[i,i]}{max(\sum_{j=1}^M \mathbf{mat}[i,j],\sum_{j=1}^M \mathbf{mat}[j,i])}}{M}
	\end{split}
\end{equation}

\begin{equation}
	\begin{split}
		G-mean = (\prod_{i=1}^M \frac{tp_i}{tp_i+fn_i})^\frac{1}{M}
	\end{split}
\end{equation}

The values of Av$F_{\beta}$, CBA and G-mean are greater, which means that the results are better.

\subsection{Statistical tests}
\label{StatisticalTest}

To obtain well-founded conclusions, the statistical studies recommended in \cite{derrac2011practical} \cite{garcia2010advanced}, are employed to compare and analyze the experimental results of contesting algorithms. Concretely, the Friedman aligned-rank test, Holm post-hoc test and Wilcoxon test are used to do the significant difference tests under performance measurements of Av$F_{\beta}$, CBA and G-mean. 

(1) Friedman aligned-ranks test \cite{hodges1962rank}. As a non-parametric statistical test, it performs multiple comparisons among the contesting methods in order to check whether they have significant differences.

To implement aligned ranks in the Friedman test, a value of location is defined as the average performance achieved by all the algorithms. Then, the difference between an algorithm's performance and the value of location can be obtained. This step is executed for each combination of algorithms and datasets, then the results are ranked from 1 to $k \cdot n$ relative to each other. This ranking scheme is the same as one that employs a multiple comparisons procedure based on independent samples\cite{kruskal1952use}. The Friedman aligned-ranks test statistic \cite{derrac2011practical} is defined as:

\begin{equation}
	\begin{split}
		F_{AR}\!\!=\!\!\frac{\left ( k-1\right )\left [ \sum_{j=1}^{k}\hat{R}_{j}^{2}-\left ( kn^{2}/4\right ){\left ( kn+1\right )}^2\right ]}{\left \{\left [ kn\left ( kn+1\right )\left ( 2kn+1\right )\right ]/6\right \}-\left ( 1/k\right )\sum_{i=1}^{n}\hat{R}_{i}^{2}}
	\end{split}
\end{equation}

\noindent where $ \hat{R}_{i} $ is equal to the rank total of the $ i$th problem and $ \hat{R}_{j} $ is the rank total of the $ j$th algorithm.To check the statistical differences among the algorithms, the test statistic $ F_{AR} $ is compared for significance with a $ \chi ^{2} $ distribution with $ k $-1 degrees of freedom. 

If the $ F_{AR} $ is greater than $ \chi ^{2} $ corresponding value, then the null hypothesis of there being no significant differences among the algorithms is rejected, i.e., they are statistically different.

(2) Holm post-hoc test. For the results of Friedman aligned-ranks test, if the significant difference exists, the Holm post-hoc test \cite{holm1979simple} is used to determine whether the best method (i.e., the control one) is statistically better than others. 

The Holm test is a one step procedure that sequentially tests the hypotheses ordered according to other methods' significance. It compares $ p_{i} $ value with $ \alpha/\left ( k-i\right ) $, starting from the method with most significant $ p $ value. If $ p_{i}$ is lower than $ \alpha/\left ( k-i\right ) $, the corresponding hypothesis is rejected, which means that it is significantly better. Otherwise, the hypothesis cannot be rejected. The test proceeds with the second, third method, and so on. 

(3) Wilcoxon signed-rank test. Regarding the comparisons between i-SVM-DE-AVG and i-SVM-DE-MAX, the Wilcoxon signed-rank test \cite{wilcoxon1945individual} is used. If $p$ value is greater than 0.05, there is no statistical difference; otherwise, it is significantly different.

\subsection{Results and discussion}
\label{Result}

Firstly, we present the experimental results based on the datasets. Tables \ref{ResGmean}, \ref{ResF1} and \ref{ResCBA} show the average results of G-mean, Av$F_{\beta}$ and CBA metrics for each method in the 15 selected datasets, and the best result for each dataset is highlighted in \textbf{bold-face}. In Table \ref{ResGmean}, SDC obtains the best average result of G-mean in 15 datasets. However, with Av$F_{\beta}$ and CBA, i-SVM-DE-AVE performs best in Tables \ref{ResF1} and \ref{ResCBA}.

\begin{table*}
	\centering
	\caption{Average results of G-mean by baseline methods and i-SVM-DE on 15 datasets. }
	\label{ResGmean}
	\scriptsize
	\resizebox{\linewidth}{!}{
	\begin{tabular}{cccccccccc}
		\hline
		Dataset & SVM            & Static-SMOTE   & Cost-SVM       & SDC            & WK-SMOTE       & PPSVM & NBSVM          & i-SVM-DE-MAX     & i-SVM-DE-AVE     \\ \hline
		Aut     & 59.87          & 63.61          & 59.98          & 62.83          & 57.70          & 52.78 & 51.05          & \textBF{70.34} & 65.43          \\
		Bal     & 96.59          & 92.18          & 97.27          & 96.54          & 87.18          & 96.93 & 92.45          & 97.25          & \textBF{97.41} \\
		Car     & 97.78          & 98.20          & 98.20          & 98.27          & 96.56          & 97.29 & 0.00           & \textBF{98.30} & 98.24          \\
		Cle     & 0.00           & 1.71           & 7.96           & 8.89           & 8.38           & 6.00  & 1.01           & 12.65          & \textBF{14.13} \\
		Der     & 96.50          & 96.51          & 96.19          & 96.58          & 95.62          & 96.54 & \textBF{97.03} & 95.03          & 94.93          \\
		Eco     & \textBF{49.89} & 49.09          & 48.75          & 45.55          & 40.27          & 48.45 & 45.22          & 46.99          & 47.32          \\
		Fla     & 35.64          & 48.28          & 54.44          & \textBF{56.28} & 47.26          & 32.08 & 42.31          & 44.11          & 37.44\\
		Gla     & 51.88          & 47.54          & 58.30          & \textBF{62.58} & 61.58          & 35.10 & 43.61          & 51.70          & 57.79          \\
		Hay     & 80.14          & 80.35          & 81.74          & 82.29          & 81.70          & 75.16 & 79.44          & 81.20          & \textBF{82.29} \\
		Hcv     & 35.33          & 38.53          & 39.57          & 35.69          & 31.73          & 24.53 & 20.64          & \textBF{41.88} & 39.77          \\
		Lym     & 66.03          & 59.69          & \textBF{76.76} & 74.10          & 62.92          & 64.35 & 55.72          & 62.16          & 62.22          \\
		New     & 93.66          & 95.31          & \textBF{97.13} & 96.48          & 96.66          & 93.00 & 86.73          & 94.98          & 94.81          \\
		Shu     & 54.77          & \textBF{55.28} & 51.01          & 54.98          & 55.05          & 48.88 & 6.85           & 54.31          & 49.80          \\
		Thy     & 68.31          & 71.32          & 84.45          & 83.55          & \textBF{85.00} & 31.06 & 21.42          & 80.05          & 81.12          \\
		Zoo     & 49.68          & 54.06          & 50.06          & 53.74          & 49.17          & 57.24 & 53.30          & \textBF{63.62} & 59.19          \\ 
		\textBF{Avg.}  & 62.40 & 63.44 & 66.79 & \textBF{67.22} & 63.79 & 57.29 & 46.45 & 66.30 & 65.46 \\
		\hline
	\end{tabular}
}
\end{table*}

\begin{table*}
	\centering
	\caption{Average results of Av$F_{\beta}$-measure by baseline methods and i-SVM-DE on 15 datasets.}
	\label{ResF1}
	\scriptsize
	\resizebox{\linewidth}{!}{
	\begin{tabular}{cccccccccc}
		\hline
		Dataset & SVM            & Static-SMOTE   & Cost-SVM       & SDC            & WK-SMOTE & PPSVM & NBSVM          & i-SVM-DE-MAX     & i-SVM-DE-AVE     \\ \hline
		Aut     & 71.82          & 73.08          & 68.46          & 72.55          & 66.45    & 69.24 & 69.87          & \textBF{76.51} & 74.58          \\
		Bal     & 94.80          & 89.56          & 95.16          & 94.29          & 83.10    & 95.77 & 87.87          & 96.60          & \textBF{96.75} \\
		Car     & 97.04          & 96.98          & 97.07          & 97.21          & 96.40    & \textBF{97.47} & 28.89 & 96.67          & 96.66          \\
		Cle     & 27.22          & 29.12          & 28.61          & 29.05          & 29.31    & 28.08 & 13.21          & \textBF{31.86} & 31.86          \\
		Der     & 96.91          & 96.90          & 96.34          & 96.81          & 95.78    & 96.93 & \textBF{97.32} & 95.42          & 95.31          \\
		Eco     & 74.73          & 70.93          & 69.82          & 63.67          & 58.90    & \textBF{75.29}         & 68.61          & 68.52          & 69.54          \\
		Fla     & 60.35          & \textBF{61.22} & 59.96          & 61.05          & 57.53    & 56.83 & 59.76          & 60.20          & 59.21          \\
		Gla     & 65.38          & 65.23          & 64.12          & 66.82          & 65.88    & 60.11 & 65.75          & 66.46          & \textBF{67.85} \\
		Hay     & 81.75          & 81.30          & 82.80          & \textBF{83.31} & 82.55    & 77.22 & 81.56          & 81.92          & 83.13          \\
		Hcv     & 66.05          & \textBF{67.69} & 66.47          & 66.18          & 59.28    & 61.17 & 18.09          & 67.14          & 67.10          \\
		Lym     & 80.67          & 77.39          & 79.73          & \textBF{81.37} & 69.98    & 79.85 & 42.42          & 78.08          & 79.16          \\
		New     & 94.50          & 95.48          & \textBF{95.64} & 95.50          & 94.43    & 94.04 & 85.46          & 95.14          & 95.12          \\
		Shu     & 87.87          & \textBF{88.98} & 83.32          & 83.77          & 68.01    & 82.57 & 32.16          & 84.02          & 83.08          \\
		Thy     & 76.60          & 78.29          & 79.23          & 79.07          & 77.62    & 45.24 & 14.78          & 82.19          & \textBF{83.23} \\
		Zoo     & 87.52          & 89.16          & 87.82          & 88.41          & 86.27    & 89.42 & 87.73          & \textBF{91.23} & 90.95          \\ 
		\textBF{Avg.}  & 77.55 & 77.42 & 76.97 & 77.27 & 72.77 & 73.95 & 56.90 & 78.13 & \textBF{78.24}  \\
		
		\hline
	\end{tabular}
}
\end{table*}

\begin{table*}[htbp]
	\centering
	\caption{Average results of CBA by baseline methods and i-SVM-DE on 15 datasets.}
	\label{ResCBA}
	\scriptsize
	\resizebox{\linewidth}{!}{
	\begin{tabular}{cccccccccc}
		\hline
		Dataset & SVM            & Static-SMOTE   & Cost-SVM       & SDC            & WK-SMOTE & PPSVM & NBSVM          & i-SVM-DE-MAX     & i-SVM-DE-AVE     \\ \hline
		Aut     & 65.08          & 66.33          & 61.59          & 66.74          & 60.18    & 62.80 & 63.07          & \textBF{69.73} & 68.33          \\
		Bal     & 92.52          & 85.84          & 92.44          & 91.08          & 77.58    & 94.16 & 82.59          & 94.88          & \textBF{95.04} \\
		Car     & 95.19          & 95.18          & 95.05          & 95.23          & 94.54    & \textBF{95.77}         & 18.52          & 94.39          & 94.36          \\
		Cle     & 25.04          & 26.52          & 24.79          & 24.69          & 24.40    & 24.78 & 10.16          & 29.05          & \textBF{29.05} \\
		Der     & 95.43          & 95.50          & 95.43          & \textBF{96.13} & 93.77    & 95.34 & 95.91          & 94.27          & 93.99          \\
		Eco     & 70.28          & 66.70          & 64.58          & 58.45          & 52.93    & \textBF{70.30}         & 64.81          & 63.34          & 64.66          \\
		Fla     & \textBF{56.85} & 56.72          & 54.60          & 56.06          & 51.95    & 53.05 & 54.41          & 54.68          & 54.31          \\
		Gla     & 60.28          & 58.30          & 55.87          & 59.51          & 58.37    & 53.71 & 58.69          & 60.68          & \textBF{61.45} \\
		Hay     & 75.60          & 76.87          & 77.03          & 75.60          & 75.07    & 72.54 & 73.47          & 76.98          & \textBF{77.77} \\
		Hcv     & 61.08          & \textBF{62.77} & 61.53          & 60.44          & 53.37    & 57.29 & 12.33          & 61.58          & 61.48          \\
		Lym     & 77.30          & 74.48          & 75.01          & \textBF{77.33} & 65.06    & 79.96 & 35.33          & 74.61          & 75.53          \\
		New     & 90.59          & \textBF{92.30} & 92.19          & 91.99          & 90.24    & 91.09 & 79.40          & 91.94          & 91.96          \\
		Shu     & 86.85          & \textBF{88.50} & 81.55          & 81.64          & 65.05    & 80.87 & 27.77          & 82.06          & 81.25          \\
		Thy     & 69.75          & 72.00          & 71.16          & 70.92          & 70.19    & 38.20 & 8.89           & 77.52          & \textBF{78.42} \\
		Zoo     & 84.71          & 87.07          & 85.48          & 86.09          & 83.39    & 86.94 & 85.41          & \textBF{90.27} & 89.60          \\ 
		\textBF{Avg.}  & 73.77 & 73.67 & 72.55 & 72.79 & 67.74 & 70.45 & 51.38 & 74.40 & \textBF{74.48} \\
		\hline
	\end{tabular}
}
\end{table*}

\begin{figure}[ht]
	\centering
	\includegraphics[width=8cm]{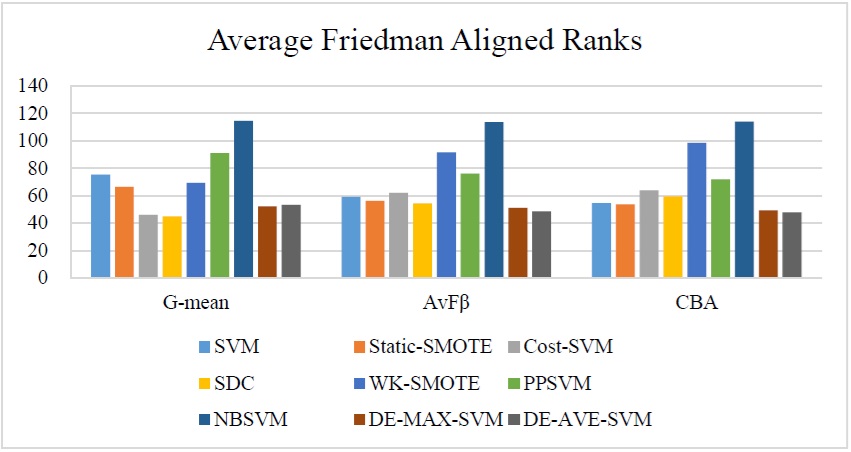}
	\caption{Average aligned rank comparisons of all the methods.}\label{fig:far}
\end{figure}

Then, the average aligned ranks of all the methods in the Friedman test are shown in Figure \ref{fig:far}. By observing it, SDC obtains the best results in terms of G-mean in average rank. For Av$F_{\beta}$ and CBA, we see that i-SVM-DE-AVE obtains the best average rank and i-SVM-DE-MAX is the second best. 

The statistical tests of Friedman aligned rank test are shown in Table \ref{tab:sta-Far}. The degree of freedom is 8, because it is calculated as $ k$-1 ($ k=9$). While degree of freedom is 8 and $ \alpha $ is 0.05, the value of $ \chi ^{2} $  is 15.50731. The results show that $ F_{AR} > \chi_{8}^{2} $ at all three metrics, which indicates that there exist significant differences among the comparison methods. Hence, we can conclude that i-SVM-DE-AVE and i-SVM-DE-MAX are significantly better than other methods at Av$F_{\beta}$ and CBA.

\begin{table}[ht]
	\centering
	\caption{Statistical test of the Friedman aligned
		rank test.}

	\begin{tabular}{p{4.235em}lll}
		\toprule
		Metric & \multicolumn{1}{p{4.235em}}{$ p $-value} & \multicolumn{1}{p{4.235em}}{$ F_{AR} $} & \multicolumn{1}{p{4.235em}}{different or not}\\
		\midrule
		G-mean & 4.82763 $\times 10^{-7} $ & 44.37740 ($ > \chi_{8}^{2} $) & {different}\\
		\multicolumn{1}{l}{Av$F_{\beta}$} & 2.78810 $\times 10^{-7}$ & 45.63536 ($ > \chi_{8}^{2} $) & {different}\\
		CBA   & 2.08476 $\times 10^{-8}$ & 51.51837 ($ > \chi_{8}^{2} $) & {different}\\
		\bottomrule
	\end{tabular}%

	\label{tab:sta-Far}%
\end{table}%

After that, we apply the Holm post-hoc test to analyze the differences between the control method and others. For G-mean, the control method is the SDC; with regard to Av$F_{\beta}$ and CBA, the control method is the i-SVM-DE-AVE. 

The statistical Holm test results with G-mean, Av$F_{\beta}$ and CBA are given in Tables \ref{tab:HoG}, \ref{tab:HoF} and \ref{tab:HoC}. Table \ref{tab:HoG} shows that it cannot reject the hypothesis with respect to i-SVM-DE-AVE and i-SVM-DE-MAX, since the corresponding $ p $ values are greater than the adjusted $ \alpha $'s. That is to say, although the results of aligned rank indicate that the SDC performs better than our proposed methods in terms of G-mean, but according to Holm test results, i-SVM-DE-AVE and i-SVM-DE-MAX remain alternatives. 

In Table \ref{tab:HoF}, we can find that the i-SVM-DE-AVE has significant differences from the NBSVM, WK-SMOTE, PPSVM, cost-SVM and SVM, with respect to Av$F_{\beta}$. We can conclude that i-SVM-DE-AVE is statistically better than these methods.

As for CBA, the significant differences exist between the i-SVM-DE-AVE and methods of NBSVM, WK-SMOTE, PPSVM, Cost-SVM and SDC, which indicate that i-SVM-DE-AVE is statistically better than these methods. Furthermore, i-SVM-DE-AVE appears to be similar with Static-SMOTE method, since there is no significant differences between them regarding to Av$F_{\beta}$ and CBA. However, in the results of Friedman test with G-mean, Av$F_{\beta}$ and CBA, i-SVM-DE-AVE has a higher rank than Static-SMOTE (shown in Fig. \ref{fig:far}) and they are statistically different (shown in Table \ref{tab:sta-Far}).


At last, the Wilcoxon tests are conducted for i-SVM-DE-AVE and i-SVM-DE-MAX, and the results are shown in Table \ref{tab:WilDE}. $ R^{+} $ corresponds to the sum of ranks for i-SVM-DE-MAX and $R^{-}$ for the i-SVM-DE-AVE. We can observe that $p$ values are greater than 0.05, which can be concluded that there is no significant difference between them with regarding to G-mean, Av$F_{\beta}$ and CBA.

\begin {comment}
From Table \ref{tab:WilG}, the Wilcoxon test shows that proposed methods achieve excellent $ p $-values in comparison with SVM, PPSVM and NBSVM, but achieves $ p $-values above the significance level by comparing with others. However, the proposed methods do not damage the performance when compared to referencing algorithms. Therefore, we can summarize that the i-SVM-DE-MAX or i-SVM-DE-AVE is very likely to offer improvements for the multi-class imbalanced classification problems.$

$By observing Table \ref{tab:WilF}, we notice that i-SVM-DE-MAX and i-SVM-DE-AVE clearly outperform the other techniques since the value of $ R^{+} $ is obviously greater than $ R^{-} $ in all comparisons. In addition, both for i-SVM-DE-MAX and i-SVM-DE-AVE, the $ p $-values observed for comparing with WK-SMOTE, PPSVM and NBSVM are lower than 0.05, which further indicates the superiority of our proposed methods.$
$The Wilcoxon tests are carried out in Table\ref{tab:WilC} when CBA is used as an indicator. In this case, all the null hypotheses of equivalence are rejected since the largest $ p $-value is equal to 0.043804, which is lower than our $ p $-value (0.05). This is much better than the case of G-mean and F1, as we observe a globally statistically significant improvement in all pairwise tests. In conclusion, it must be noted that the proposed methods are worthwhile choice for such problems.$
\end {comment}

\begin{table}
	\centering
	\caption{Holm test of G-mean}
	\begin{tabular}{lllllp{6.5em}}
		\toprule
		\multicolumn{5}{p{26.005em}}{Control method: SDC} \\
		\midrule
		i     & Algorithm & $ p $-value & $ \alpha/i$  & \multicolumn{1}{l}{Hypothesis($ \alpha$=0.05)} & {better or not}\\
		8     & NBSVM & 0     & 0.00625 & \multicolumn{1}{l}{Rejected} & {better}\\
		7     & PPSVM & 0     & 0.00714 & \multicolumn{1}{l}{Rejected} & {better}\\
		6     & SVM   & 6.21176$ \times 10^{-10} $ & 0.00833 & \multicolumn{1}{l}{Rejected} & {better} \\
		5     & WK-SMOTE & 7.04146$ \times 10^{-7} $ & 0.01  & \multicolumn{1}{l}{Rejected} & {better} \\
		4     & Static-SMOTE & 0.00001 & 0.0125 & \multicolumn{1}{l}{Rejected} & {better}\\
		3     & i-SVM-DE-AVE & 0.08641 & 0.01667 & \multicolumn{1}{l}{Not rejected} & {not}\\
		2     & i-SVM-DE-MAX & 0.13981 & 0.025 & \multicolumn{1}{l}{Not rejected} & {not}\\
		1 & Cost-SVM & 0.81705 & 0.05  & Not rejected & {not}\\
		\bottomrule
	\end{tabular}%
	\label{tab:HoG}%
\end{table}%

\begin{table}
	\centering
	\caption{Holm test of Av$F_{\beta}$}
	\begin{tabular}{lllllp{6.5em}}
		\toprule
		\multicolumn{5}{p{26.77em}}{Control method: i-SVM-DE-AVE} \\
		\midrule
		i     & \multicolumn{1}{l}{Algorithm} & $ p $-value & $ \alpha/i$  & \multicolumn{1}{l}{Hypothesis($ \alpha$=0.05)} & {better or not}\\
		8     & \multicolumn{1}{l}{NBSVM} & 0     & 0.00625 & \multicolumn{1}{l}{Rejected} &{better}\\
		7     & \multicolumn{1}{l}{WK-SMOTE} & 0     & 0.00714 & \multicolumn{1}{l}{Rejected} &{better}\\
		6     & \multicolumn{1}{l}{PPSVM} & 2.14592$ \times 10^{-8} $ & 0.00833 & \multicolumn{1}{l}{Rejected} &{better}\\
		5     & \multicolumn{1}{l}{Cost-SVM} & 0.00586 & 0.01  & \multicolumn{1}{l}{Rejected} &{better}\\
		4     & \multicolumn{1}{l}{SVM} & 0.03434 & 0.0125 & \multicolumn{1}{l}{Rejected} &{better}\\
		3     & \multicolumn{1}{l}{Static-SMOTE} & 0.12246 & 0.01667 & \multicolumn{1}{l}{Not rejected} &{not}\\
		2     & \multicolumn{1}{l}{SDC} & 0.23376 & 0.025 & \multicolumn{1}{l}{Not rejected} &{not}\\
		1 & i-SVM-DE-MAX & 0.61461 & 0.05  & Not rejected &{not}\\
		\bottomrule
	\end{tabular}%
	\label{tab:HoF}%
\end{table}%

\begin{table}[htbp]
	\centering
	\caption{Holm test of CBA}
	\begin{tabular}{lllllp{6.5em}}
		\toprule
		\multicolumn{5}{p{26.77em}}{Control method: i-SVM-DE-AVE} \\
		\midrule
		i     & \multicolumn{1}{l}{Algorithm} & $ p $-value & $\alpha/i$  & \multicolumn{1}{l}{Hypothesis($ \alpha$=0.05)} & {better or not} \\
		8     & \multicolumn{1}{l}{NBSVM} & 0     & 0.00625 & \multicolumn{1}{l}{Rejected} &{better}\\
		7     & \multicolumn{1}{l}{WK-SMOTE} & 0     & 0.00714 & \multicolumn{1}{l}{Rejected} &{better}\\
		6     & \multicolumn{1}{l}{PPSVM} & 1$ \times 10^{-6} $  & 0.00833 & \multicolumn{1}{l}{Rejected} &{better}\\
		5     & \multicolumn{1}{l}{Cost-SVM} & 0.00117 & 0.01  & \multicolumn{1}{l}{Rejected} &{better}\\
		4     & \multicolumn{1}{l}{SDC} & 0.01822 & 0.0125 & \multicolumn{1}{l}{Rejected} &{better}\\
		3     & \multicolumn{1}{l}{SVM} & 0.17357 & 0.01667 & \multicolumn{1}{l}{Not rejected} &{not}\\
		2     & \multicolumn{1}{l}{Static-SMOTE} & 0.23915 & 0.025 & \multicolumn{1}{l}{Not rejected} &{not}\\
		1 & i-SVM-DE-MAX & 0.79073 & 0.05  & Not rejected &{not}\\
		\bottomrule
	\end{tabular}%
	\label{tab:HoC}%
\end{table}%

\begin {comment}
\begin{table*}[htbp]
	\centering
	\caption{Results of Wilcoxon tests for comparison of proposed method and the state-of-the-art methods for G-mean. A ``$ \ast $" near the $ p $-value means that there are statistical differences with $ \alpha $ = 0.05(95$ \% $ confidence) and a ``$ - $" with $ \alpha $ = 0.1 (90$ \% $ confidence).}
	\begin{tabular}{llllllll}
		\toprule
		Comparison & $ R^{+} $     & $ R^{-} $     & $ p $-value & Comparison & $ R^{+} $     & $ R^{-} $     & $ p $-value \\
		\midrule
		i-SVM-DE-MAX vs. SVM & 93    & 27    & 0.06372$ - $ & i-SVM-DE-AVE vs. SVM & 91    & 29    & 0.08325$ - $ \\
		i-SVM-DE-MAX vs. Static-SMOTE & 93    & 27    & 0.06372$ - $ & i-SVM-DE-AVE vs. Static-SMOTE & 84    & 36    & 0.18762 \\
		i-SVM-DE-MAX vs. Cost-SVM & 54    & 66    & 0.76154 & i-SVM-DE-AVE vs. Cost-SVM & 47    & 73    & 0.48871 \\
		i-SVM-DE-MAX vs. SDC & 51    & 69    & 0.63867 & i-SVM-DE-AVE vs. SDC & 45    & 60    & 0.63777 \\
		i-SVM-DE-MAX vs. WK-SMOTE & 78    & 42    & 0.33026 & i-SVM-DE-AVE vs. WK-SMOTE & 76    & 44    & 0.38940\\
		i-SVM-DE-MAX vs. PPSVM & 107   & 13    & 0.00537$ \ast $ & i-SVM-DE-AVE vs. PPSVM & 103   & 17    & 0.01245$ \ast $ \\
		i-SVM-DE-MAX vs. NBSVM & 116   & 4     & 0.00043$ \ast $ & i-SVM-DE-AVE vs. NBSVM & 115   & 5     & 0.00061$ \ast $ \\
		\bottomrule
	\end{tabular}%
	\label{tab:WilG}%
\end{table*}%

$\begin{table*}[htbp]
	\centering
	\caption{Results of Wilcoxon tests for comparison of proposed method and the state-of-the-art methods for F1. A ``$ \ast $" near the $ p $-value means that there are statistical differences with $ \alpha $ = 0.05(95$ \% $ confidence) and a ``$ - $" with $ \alpha $ = 0.1 (90$ \% $ confidence).}
	\begin{tabular}{llllllll}
		\toprule
		Comparison & $ R^{+} $     & $ R^{-} $     & $ p $-value & Comparison & $ R^{+} $     & $ R^{-} $     & $ p $-value \\
		\midrule
		i-SVM-DE-MAX vs. SVM & 74    & 46    & 0.45428 & i-SVM-DE-AVE vs. SVM & 75    & 45    & 0.42120 \\
		i-SVM-DE-MAX vs. Static-SMOTE & 76    & 44    & 0.38940 & i-SVM-DE-AVE vs. Static-SMOTE & 80    & 40    & 0.27686 \\
		i-SVM-DE-MAX vs. Cost-SVM & 84    & 36    & 0.18762 & i-SVM-DE-AVE vs. Cost-SVM & 85    & 35    & 0.16882 \\
		i-SVM-DE-MAX vs. SDC & 78    & 42    & 0.33026 & i-SVM-DE-AVE vs. SDC & 85    & 35    & 0.16882 \\
		i-SVM-DE-MAX vs. WK-SMOTE & 114   & 6     & 0.00085$ \ast $ & i-SVM-DE-AVE vs. WK-SMOTE & 118   & 2     & 0.00018$ \ast $ \\
		i-SVM-DE-MAX vs. PPSVM & 95    & 25    & 0.04791$ \ast $ & i-SVM-DE-AVE vs. PPSVM & 97    & 23    & 0.03534$ \ast $ \\
		i-SVM-DE-MAX vs. NBSVM & 114   & 6     & 0.00085$ \ast $ & i-SVM-DE-AVE vs. NBSVM & 115   & 5     & 0.00061$ \ast $ \\
		\bottomrule
	\end{tabular}%
	\label{tab:WilF}%
\end{table*}%

$\begin{table*}[htbp]
	\centering
	\caption{Results of Wilcoxon tests for comparison of proposed method and the state-of-the-art methods for CBA. A ``$ \ast $" near the $ p $-value means that there are statistical differences with $ \alpha $ = 0.05(95$ \% $ confidence) and a ``$ - $" with $ \alpha $ = 0.1 (90$ \% $ confidence).}
	\begin{tabular}{llllllll}
		\toprule
		Comparison & $ R^{+} $     & $ R^{-} $     & $ p $-value & Comparison & $ R^{+} $     & $ R^{-} $     & $ p $-value \\
		\midrule
		i-SVM-DE-MAX vs. SVM & 110   & 10    & 0.00262$ \ast $ & i-SVM-DE-AVE vs. SVM & 111   & 9     & 0.00201$ \ast $ \\
		i-SVM-DE-MAX vs. Static-SMOTE & 110   & 10    & 0.00262$ \ast $ & i-SVM-DE-AVE vs. Static-SMOTE & 109   & 11    & 0.00336$ \ast $ \\
		i-SVM-DE-MAX vs. Cost-SVM & 119   & 1     & 0.00012$ \ast $ & i-SVM-DE-AVE vs. Cost-SVM & 119   & 1     & 0.00012$ \ast $ \\
		i-SVM-DE-MAX vs. SDC & 119   & 1     & 0.00012$ \ast $ & i-SVM-DE-AVE vs. SDC & 119   & 1     & 0.00012$ \ast $ \\
		i-SVM-DE-MAX vs. WK-SMOTE & 120   & 0     & 0.00006$ \ast $ & i-SVM-DE-AVE vs. WK-SMOTE & 120   & 0     & 0.00006$ \ast $ \\
		i-SVM-DE-MAX vs. PPSVM & 113   & 7     & 0.00116$ \ast $ & i-SVM-DE-AVE vs. PPSVM & 114   & 6     & 0.00085$ \ast $ \\
		i-SVM-DE-MAX vs. NBSVM & 119   & 1     & 0.00012$ \ast $ & i-SVM-DE-AVE vs. NBSVM & 119   & 1     & 0.00012$ \ast $ \\
		\bottomrule
	\end{tabular}%
	\label{tab:WilC}%
\end{table*}%
\end{comment}

\begin{table*}[htbp]
\centering
\caption{Wilcoxon test results of comparisons between i-SVM-DE-MAX and i-SVM-DE-AVE. }
\begin{tabular}{llllll}
	\toprule
	Measure & Comparison & $ R^{+} $     & $ R^{-} $     & $ p $-value & {different or not}\\
	\midrule
	G-mean & i-SVM-DE-MAX vs. i-SVM-DE-AVE & 70    & 49.5  & 0.59949 &{not}\\
	Av$F_{\beta}$    & i-SVM-DE-MAX vs. i-SVM-DE-AVE & 45    & 60    & 0.63777 &{not}\\
	CBA   & i-SVM-DE-MAX vs. i-SVM-DE-AVE & 47    & 58    & 0.72989 &{not}\\
	\bottomrule
\end{tabular}%
\label{tab:WilDE}%
\end{table*}%

\section{Conclusions}\label{conclusion}

In this paper, we propose the i-SVM-DE methods to deal with the multi-class imbalanced classification tasks. An i-SVM model is proposed to handle the data imbalance problems. Comparing to the classical SVM, the modification of constraints make i-SVM be more generalized. The multi-class problem is decomposed into binary subproblems by using the OVO strategy. Eventually, a multi-class SVM with data imbalance problem is reformulated to a large parameter optimization problem. An improved DE algorithm is applied to obtain the support vectors for each class simultaneously. It is notable that the existing studies of combining evolutionary algorithms into machine learning models need the validation sets to evaluate the learned model and tune hyper-parameters. In this paper, we propose the fitness functions to evaluate the individuals and guide the search directions, and the obtained optimal parameters construct the final classification model. This method has great advantages in handling small-sample data, since the datasets are split into training and testing sets.

Based on the proposed fitness functions, the corresponding methods are called i-SVM-DE-AVE and i-SVM-DE-MAX, respectively.
A thorough set of experiments are carried out on 15 selected datasets. The metrics of G-mean,  Av$F_{\beta}$ and CBA are employed to evaluate the performances of algorithms. The results show that i-SVM-DE-AVE has a higher rank in the Friedman test at the metrics of Av$F_{\beta}$ and CBA. When it is considered as the control method, it is significantly different from most of the baseline methods in the Holm test. With regard to G-mean, the SDC performs best in the Friedman rank test. But, according to Holm test, SDC is not significantly different from i-SVM-DE-AVE and i-SVM-DE-MAX. We can conclude that our proposed method has its benefits impressively.   

While the optimal parameters are determined, i-SVM-DE predict the class labels of unseen samples in the same way as classical SVM. Due to the DE algorithm used in the training phase, this method may be time-consuming, and improving the model's efficiency will be our future works.

\section{Acknowledgments}\label{supportinginformation}

This work was supported in part by the National Natural Science Foundation of China (NSFC) (Grant Nos. 72171065 and 71831006).






\end{document}